\documentclass[letterpaper]{article} 
\usepackage{aaai23}  
\usepackage{times}  
\usepackage{helvet}  
\usepackage{courier}  
\usepackage[hyphens]{url}  
\usepackage{graphicx} 
\usepackage{natbib}  
\usepackage{caption} 
\usepackage{algorithm}
\usepackage{algorithmic}

\usepackage{newfloat}
\usepackage{listings}
\DeclareCaptionStyle{ruled}{labelfont=normalfont,labelsep=colon,strut=off} 
\floatstyle{ruled}
\newfloat{listing}{tb}{lst}{}
\floatname{listing}{Listing}
\usepackage{mathtools}
\usepackage{graphicx}
\usepackage{multirow}
\graphicspath{{Figures/}}
\usepackage{array}
\usepackage{amssymb}
\usepackage{epstopdf}
\usepackage{array}
\usepackage{adjustbox}
\usepackage{subfigure}
\usepackage{booktabs}
\usepackage{diagbox}
\usepackage{amsfonts} 
\usepackage{dsfont}  

\title{MHCCL: Masked Hierarchical Cluster-Wise Contrastive Learning for Multivariate Time Series}

\author{
    Qianwen Meng\textsuperscript{\rm 1,\rm 2},
    Hangwei Qian\textsuperscript{\rm 3}\thanks{Corresponding author},
    Yong Liu\textsuperscript{\rm 4},
    Lizhen Cui\textsuperscript{\rm 1,\rm 2}$^*$,
    Yonghui Xu\textsuperscript{\rm 1,\rm 2}$^*$,
    Zhiqi Shen\textsuperscript{\rm 4}
}
\affiliations{
    \textsuperscript{\rm 1}School of Software, Shandong University, Jinan, China\\
    \textsuperscript{\rm 2}Joint SDU-NTU Centre for Artificial Intelligence Research (C-FAIR), Shandong University, Jinan, China\\
    \textsuperscript{\rm 3}Lund University, Sweden\\
    \textsuperscript{\rm 4}School of Computer Science and Engineering, Nanyang Technological University, Singapore\\
    mqw\_sdu@mail.sdu.edu.cn, hangwei.qian@math.lth.se, stephenliu@ntu.edu.sg, clz@sdu.edu.cn, xuyonghui@sdu.edu.cn, zqshen@ntu.edu.sg
}

\begin{document}

\maketitle

\begin{abstract}
Learning semantic-rich representations from raw unlabeled time series data is critical for downstream tasks such as classification and forecasting. Contrastive learning has recently shown its promising representation learning capability in the absence of expert annotations. However, existing contrastive approaches generally treat each instance independently, which leads to false negative pairs that share the same semantics. To tackle this problem, we propose MHCCL, a \underline{M}asked \underline{H}ierarchical \underline{C}luster-wise \underline{C}ontrastive \underline{L}earning model, which exploits semantic information obtained from the hierarchical structure consisting of multiple latent partitions for multivariate time series. Motivated by the observation that fine-grained clustering preserves higher purity while coarse-grained one reflects higher-level semantics, we propose a novel downward masking strategy to filter out fake negatives and supplement positives by incorporating the multi-granularity information from the clustering hierarchy. In addition, a novel upward masking strategy is designed in MHCCL to remove outliers of clusters at each partition to refine prototypes, which helps speed up the hierarchical clustering process and improves the clustering quality. We conduct experimental evaluations on seven widely-used multivariate time series datasets. The results demonstrate the superiority of MHCCL over the state-of-the-art approaches for unsupervised time series representation learning. 
\end{abstract}

\section{Introduction}

Multivariate time series data are ubiquitous in various domains such as industry, finance and healthcare~\cite{dtw,tssurvey,tsae,trd}. Such data record the changing trends of multiple variables over time. Time series representation learning is a way to transform the complex raw time series data into semantic-rich representations~\cite{tstcc,review_temporal}. Since time series data are usually collected through a large number of sensors or devices, there are no obvious visual patterns that can be easily recognized by humans. As a result, extensive efforts on data annotations are required before supervised learning approaches can be performed~\cite{DBLP:journals/ai/QianPM21}. Such an expensive manual labeling process greatly limits the utilization of time series data. Therefore, unsupervised learning has emerged and gained the favor of researchers. There are unsupervised learning methods for time series that adopt auto-encoders~\cite{timenet} and seq-to-seq models~\cite{icp} to reconstruct the raw time series through joint training with decoders. However, robust reconstruction of complex time series is challenging in many cases, especially for the high-frequency physiological signals~\cite{tnc,sslecg}. In this case, self-supervised learning which uses the self-generated supervised signals obtained by pretext tasks has been developed as well~\cite{ssl1,ssl2}.

In this paper, we focus on contrastive learning, which is a special form of self-supervised learning with instance discrimination as the pretext task. Contrastive learning has achieved remarkable advantages in diverse applications such as image~\cite{simclr,byol,moco,respnet,tclr} and time series classification~\cite{tloss,tstcc,ts2vec,evaluation_clf_17}. Nevertheless, the aforementioned approaches mainly focus on instance-level contrast that treats instances independently. They usually regard the augmented view of the same instance as the unique positive and the remaining views as negatives. Then, instances are distinguished by pulling representations of the positive pair together and pushing those of negative pairs apart. As the unique positive pair is composed of augmented views generated from the same instance, other instances with similar higher-level implicit semantics are misjudged as negative ones. In this case, the false negative instances will be pushed away in subsequent contrastive learning, which causes an adverse effect on instance discrimination.

Recently, cluster-wise contrastive learning breaks the independence between instances by exploiting the latent cluster information among instances. The learned representations are expected to retain higher-level semantic information by taking advantage of additional prior information brought by clustering~\cite{pcl}. However, existing cluster-wise contrastive learning approaches~\cite{swav,pcl,c3gan} usually i) adopt flat clustering algorithms that only capture a single hierarchy of semantic clusters, and ii) require prior knowledge to pre-specify the number of clusters, which are non-trivial for unlabeled time series. What's worse, these approaches still suffer from fake negatives and limited positives, as they construct only one positive pair with its corresponding prototype, while treating all remaining prototypes as negative candidates. Such clustering is prone to noise and fails to fully take advantage of the hierarchical semantic information behind the entire set of instances. Without ground-truth labels, it is challenging to guarantee the accurate constructions of such positive and negative pairs. For example, two instances may be divided into different clusters at a fine-grained level, but belong to the same cluster at a coarse-grained level. The fine-grained prototypes preserve higher purity, while the coarse-grained ones reflect higher-level semantics. Therefore, the implicit hierarchical semantics are all valuable and should be taken into account.

In this paper, we propose a \underline{M}asked \underline{H}ierarchical \underline{C}luster-wise \underline{C}ontrastive \underline{L}earning (MHCCL)\footnote{Code is available at \url{https://github.com/mqwfrog/MHCCL}.} model for time series representation learning. MHCCL is facilitated with hierarchical clustering to enable more informative positive and negative pairs. It is motivated by the observation of multi-granularity of clustering, i.e., clustering with a larger number of clusters preserves high purity within each small cluster, while the one with a smaller number of clusters can better reflect high-level semantics. We propose novel downward and upward masking strategies to improve constructed pairs for multi-level contrast, which is achieved by incorporating information from the remaining hierarchy at each partition of clustering. Downward masking utilizes the information from upper partitions to lower partitions in the hierarchy, which helps supplement latent positive pairs and filter out fake negative pairs for effective contrastive learning. In addition, upward masking utilizes the information from lower partitions to upper partitions, which helps remove outliers and refine prototypes to improve the clustering quality.

The main contributions of this work are as follows. \textit{Firstly}, we propose a novel downward masking strategy to incorporate the implicit multi-granularity semantic information obtained by hierarchical clustering into the construction of positive and negative pairs for contrastive learning. \textit{Secondly}, we reserve the representative instances that can characterize the cluster adequately, and filter the outliers that may bring side effects on contrastive learning by a novel upward masking strategy during hierarchical clustering. \textit{Thirdly}, we conduct extensive experiments to evaluate the proposed MHCCL on seven benchmark datasets, with results demonstrating the effectiveness of MHCCL.

\section{Related Work}

\paragraph{Instance-Wise Contrastive Learning.}
These methods treat each instance independently and design the instance discrimination pretext task to keep similar instances close and dissimilar instances far away~\cite{DBLP:conf/kdd/QianTM22}. They construct positive and negative pairs mainly based on the augmented representations of original instances in the embedding space. SimCLR~\cite{simclr} treats augmented views of the same instance as the unique positive pair, and all remaining ones within the minibatch as negative pairs. MoCo~\cite{moco} designs the momentum encoder and a dynamic queue to maintain more negative candidates across minibatches. There also exist approaches such as BYOL~\cite{byol} that totally avoid negative pairs by facilitating the predictor with the stop-gradient strategy. The aforementioned approaches demonstrate significant advantages on image data, but may not work well on time series. Therefore, researchers also improve contrastive learning methods for time series by taking the temporal dependency into account. T-Loss~\cite{tloss} employs an efficient triplet loss that uses time-based negative sampling to distinguish anchors from negative instances, and assimilate anchors and positive instances. ~\citeauthor{tstcc}~\shortcite{tstcc} proposed TS-TCC to learn time series representations with cross-view temporal and contextual contrasting. ~\citeauthor{ts2vec}~\shortcite{ts2vec} presented TS2Vec which focused on the hierarchy within the same time series, to distinguish the multi-scale contextual information at both timestamp-level and instance-level. However, the methods mentioned above are unable to capture higher-level semantic structures involved in data, which leads to false negatives.

\paragraph{Cluster-Wise Contrastive Learning.} 
These methods alleviate the proportion of false negatives that share similar higher-level semantics~\cite{crosslevel}. CC~\cite{cc} utilizes the feature matrix whose dimensionality is equal to the number of clusters to obtain prototypes. However, it requires prior knowledge to pre-specify the number of clusters, which is non-trivial for unlabeled data. SwAV~\cite{swav} performs online clustering to obtain prototypes and enables clustering-based discrimination between groups of instances with similar features instead of individuals. PCL~\cite{pcl} estimates the distribution of prototypes via clustering in E-step and optimizes the network by contrastive learning in M-step iteratively. All three methods adopt flat clustering that only captures a single hierarchy of semantics, and the number of clusters needs to be manually specified, which could not capture the natural hierarchies in data. In contrast, hierarchical clustering is more advantageous in capturing multiple levels of granularity in the underlying data structure. Recent studies~\cite{ccl,slic} are most related to our work, as they also explore hierarchical clustering during contrastive learning. However, they only use the information at a specific clustering partition, while our MHCCL fully utilizes the information of multiple partitions in the hierarchical structure for multi-level contrast. 

\section{The Proposed MHCCL}

\begin{figure*} 
\centering 
\includegraphics{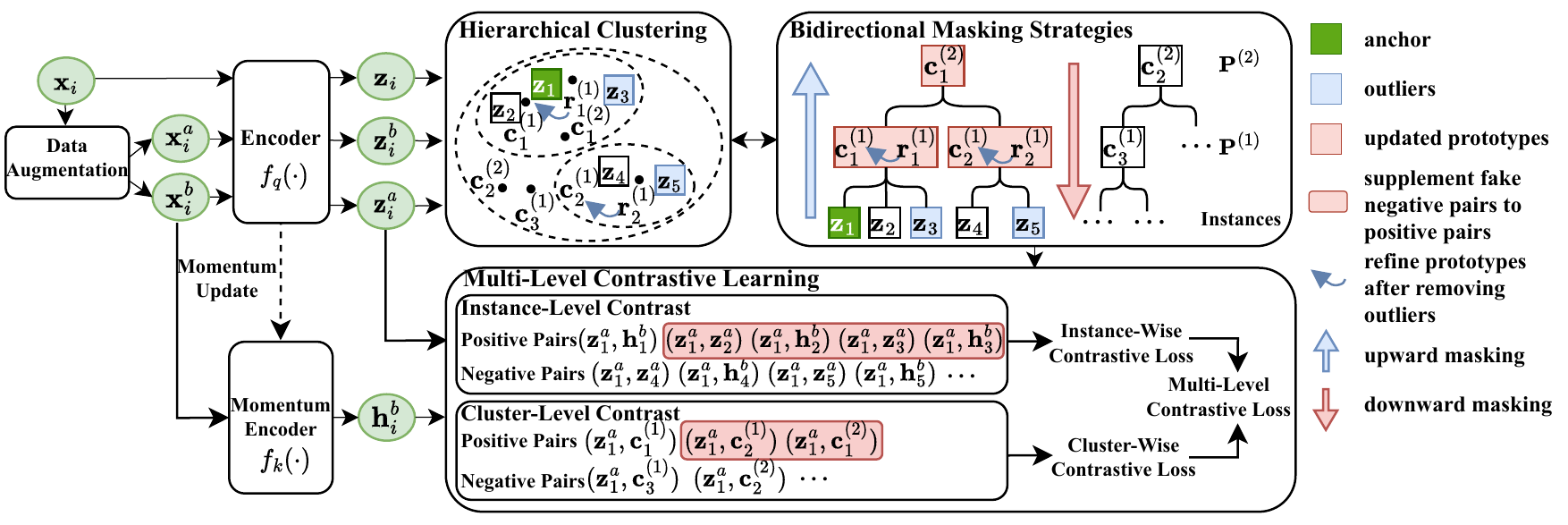}
\caption{Overview of the proposed MHCCL model. Firstly, an encoder maps two augmented views produced by the data augmentation module into latent vectors respectively. Meanwhile, a momentum encoder generates another view for multi-level contrast. Next, the hierarchical clustering module generates multiple partitions, and upward masking is conducted iteratively with hierarchical clustering to remove outliers and refine prototypes, which helps improve the clustering quality. Then, downward masking filters out fake negatives and supplements positives for multi-level contrast. Finally, the multi-level contrastive learning module uses the constructed pairs to perform training by summing up instance-wise and cluster-wise losses.}
 \label{fig:overview}
\end{figure*}

\paragraph{Problem Definition.}
Given a set of $N$ multivariate time series $X=\{\mathbf{x}_i\}_{i=1}^N$, each time series $\mathbf{x}_i \in \mathbb{R} ^ {T \times V }$ is a sequence composed of $V$ variables within $T$ timestamps. The goal of unsupervised representation learning is to train an encoder $f_q(\cdot)$, which maps each raw time series $\mathbf{x}_i$ into $\mathbf{z}_i=f_q(\mathbf{x}_i) \in \mathbb{R}^{D\times 1}$, such that the learned representations $\{\mathbf{z}_i\}_{i=1}^N$ are beneficial for downstream classification tasks. MHCCL achieves this goal by optimizing a contrastive loss

\begin{equation}
\mathcal{L}^{(i)} =-\log  \frac{ \sum_{k=1}^{S_{\text{pos}}}\exp (\text{sim}(\mathbf{z}_{i}^{a},\mathbf{h}_{k}^{b})/\tau) }
{\sum_{j=1}^{S_{\text{\text{neg}}}} \exp (\text{sim}(\mathbf{z}_{i}^{a},\mathbf{h}_{j}^{b})/\tau)}, 
\label{infonce}
\end{equation}
where $\text{sim}(\cdot,\cdot)$ denotes the similarity metric implemented by cosine similarity, $\tau$ is the temperature coefficient, $S_{\text{pos}}$ and $S_{\text{neg}}$ denote the number of positives  and negatives. Eq.~\eqref{infonce} is a variant of InfoNCE with the difference of varying sets of positives, and it becomes the standard InfoNCE when $S_{\text{pos}}=1$. For implementation, it can be rewritten as $-\sum_{k=1}^{S_{\text{pos}}}\log(\sigma(\text{sim}(\mathbf{z}_{i}^{a},\mathbf{h}_{k}^{b}))))-\sum_{j=1}^{S_{\text{neg}}}\log(\sigma(-\text{sim}(\mathbf{z}_{i}^{a},\mathbf{h}_{j}^{b})))$.

\paragraph{MHCCL Overview.}
As illustrated in Figure~\ref{fig:overview}, for each input data $\mathbf{x}_{i}$, the data augmentation module produces two augmented views $\mathbf{x}_i^a$ and $\mathbf{x}_i^b$. Then the encoder $f_{q}(\cdot)$ maps 
$\mathbf{x}_i$, $\mathbf{x}_i^a$ and $\mathbf{x}_i^b$ 
into latent vectors $\mathbf{z}_i=f_q(\mathbf{x}_i)$, $\mathbf{z}_i^a=f_q(\mathbf{x}_i^a)$, $\mathbf{z}_i^b=f_q(\mathbf{x}_i^b)$, respectively. Meanwhile, a momentum encoder $f_{k}(\cdot)$ generates another view $\mathbf{h}_i^b=f_k(\mathbf{x}_i^b)$. $f_{q}$'s parameter $\theta_{q}$ is updated by stochastic gradient descent with back-propagation, while $f_{k}$'s $\theta_{k}$ is updated by the momentum-based moving average to maintain consistency, i.e., $\theta_{k}\leftarrow m\theta_{k}+(1-m)\theta_{q}$ with $m\in[0,1)$ denoting the momentum coefficient. The hierarchical clustering module generates $M$ partitions, where $\text{P}^{(p)}$ denotes the $p$-th partition, and each partition contains $K_p$ clusters. We use $M=2$ for illustration, where $\text{P}^{(1)}$ and $\text{P}^{(2)}$ are named from bottom to top. The core of effective contrastive learning is the construction of positive and negative pairs, and MHCCL introduces two novel masking strategies to improve the quality of pairs. Specifically, upward masking is conducted iteratively with hierarchical clustering to help remove outliers and refine the prototypes. Then, downward masking helps filter out false negatives and construct better pairs for multi-level contrast. Finally, the instance-wise and cluster-wise losses are summed up for training.

\paragraph{Hierarchical Clustering.}
According to the example given in Fig.~\ref{fig:overview}, the original prototype $\mathbf{r}_{2}^{(1)}$ is misjudged as the false negative candidate of anchor $\mathbf{z}_1$ in flat clustering that only captures a single hierarchy of semantic clusters. However, it can be observed that $\mathbf{r}_{2}^{(1)}$ and $\mathbf{r}_{1}^{(1)}$ belong to the same cluster at next partition $\text{P}^{(2)}$ from the latent higher-level hierarchical clustering results. Such false pairs have negative effects on contrastive learning. Intuitively, we envision that if the false pairs can be recognized and filtered out in advance, the results are supposed to be improved. However, since time series data are unlabeled during the unsupervised learning process, the correctness of clustering results cannot be guaranteed. In this case, motivated by the observation that hierarchical clustering can discover clusters at different levels of granularity, we take the entire natural hierarchical structure into consideration. It is worth noting that the bottom fine-grained partition preserves higher purity while the top coarse-grained one reflects higher-level semantics, both of which are important for measuring similarity between instances. To this end, MHCCL effectively utilizes the hierarchical clustering algorithm to capture different grained groupings and generate hierarchical prototypes. We hope to utilize the purity information at different levels of clustering and the latent correlations among multiple partitions in the hierarchical structure to reduce false pairs. Our proposed approach is very flexible as it has no restrictions on the choice of hierarchical clustering algorithms. In this paper, we adopt the FINCH~\cite{finch} algorithm which utilizes the first nearest neighbor to construct the clustering hierarchy in a bottom-up manner. To be specific, MHCCL first computes the euclidean distances between all representations of raw time series and their augmented views to find the nearest neighbor $\mathbf{\omega}_i^1$ for each time series $\mathbf{x}_i$. Assume that $\mathbf{x}_j$ is the nearest neighbor of time series $\mathbf{x}_i$, it can be denoted as $\mathbf{\omega}_i^1=j$. The adjacency link matrix $\mathbf{A}(i,j)$ is then obtained by assigning $\mathbf{A}(i,j) = 1$ if $\mathbf{\omega}_i^1=j ~\text{or}~ \mathbf{\omega}_j^1=i ~\text{or}~ \mathbf{\omega}_i^1=\mathbf{\omega}_j^1$ and 0 otherwise.
With the help of $\mathbf{A}$, the first clustering partition is formed, and the prototype of each cluster is calculated as the mean of instances within the cluster. The clustering process will stop when there are less than two clusters. We notice that CCL~\cite{ccl} proposes a contrastive model based on FINCH as well, yet it lacks the construction and selection of contrastive pairs using multi-level prototypes as MHCCL does.

\subsection{Instance-Level Contrast}
Recall that popular instance-wise contrastive methods~\cite{simclr,tstcc} construct only 1 positive pair and $2(B-1)$ negative pairs within a minibatch of size $B$. It is unreasonable that all remaining instances are treated as negative candidates. Intuitively, instances belonging to the same class at a high-purity clustering partition have higher similarity, which should not be negatives but instead be treated as positives. Therefore, MHCCL proposes to supplement positive candidates belonging to the same cluster at the bottom partition $\text{P}^{(1)}$ within the minibatch. MHCCL randomly selects $S_{\text{pos}}-1$ instances to form positive pairs with the anchor $\mathbf{z}_i$ in addition to $\mathbf{h}_i^b$ encoded by the momentum encoder. As shown in Fig.~\ref{fig:overview}, false negatives misjudged by existing instance-level approaches are supplemented to be positives by MHCCL, which are marked with red boxes. To be specific, $\mathbf{z}_2$ and $\mathbf{z}_3$ are treated as positive candidates because they belong to the same cluster as the anchor $\mathbf{z}_1$ at partition $\text{P}^{(1)}$. On the contrary, remaining instances such as $\mathbf{z}_4$ and $\mathbf{z}_5$ are all treated as negative candidates. For the convenience of implementation, if there are less than $S_{\text{pos}}-1$ positive candidates of the anchor within the minibatch, $\mathbf{h}_i^b$ will be replicated to fill in the insufficient positive pairs. Meanwhile, MHCCL randomly selects $S_{\text{neg}}$ negative instances from remaining instances, to form negative pairs with the anchor. Since each positive or negative candidate has two augmented views, there are in total $2 \times S_{\text{pos}}$ positive and $2 \times S_{\text{neg}}$ negative instance-wise pairs. 

\subsection{Cluster-Level Contrast}

\paragraph{Upward Masking Strategy in Hierarchical Clustering.} 

In hierarchical clustering, prototypes (i.e. cluster centroids) at the bottom partition are typically obtained by computing the mean vector within each cluster of instances, while prototypes at subsequent partitions are calculated according to the mean of prototypes at the previous partition. So, there is a strong dependency between multiple partitions. Also, prototypes are sensitive to outliers at each partition, which in turn affects the clustering accuracy of subsequent partitions. As such, the outliers should be filtered. Considering the above characteristics, MHCCL proposes the upward masking strategy to update the upper structure by performing masking in the lower structure of the hierarchical tree. In other words, MHCCL removes outliers and updates prototypes at each partition from bottom to top during hierarchical clustering. Note that the strategy of removing outliers can be flexible. Here we adopt a threshold-based strategy, i.e., instances whose distance from the prototype is above a specified threshold will be masked. There also exist other strategies such as masking a fixed proportion of outliers, which will be compared in Experiments. In this way, after each partition of clustering, MHCCL first calculates mean prototypes at the current partition, and then removes outliers based on the threshold through upward masking. Next, mean prototypes will be refined to better reflect the resultant clusters, which are better candidates for the construction of positives and negatives. Intuitively, removing outliers will help improve the quality of hierarchical clustering, and the recalculated prototypes used for cluster-wise contrastive learning will be more accurate. As illustrated in Figure~\ref{fig:overview}, during hierarchical clustering, upward masking is adopted to remove outliers including $\mathbf{z}_3$ and $\mathbf{z}_5$, while updating original prototypes $\mathbf{r}_{1}^{(1)}$, $\mathbf{r}_{2}^{(1)}$, $\mathbf{r}_{1}^{(2)}$ to $\mathbf{c}_{1}^{(1)}$, $\mathbf{c}_{2}^{(1)}$, $\mathbf{c}_{1}^{(2)}$ from lower partitions to upper partitions. For illustrative simplicity, other original prototypes are omitted other than $\mathbf{r}_{1}^{(1)}$ and $\mathbf{r}_{2}^{(1)}$.

\paragraph{Downward Masking Strategy in Contrastive Pairs Selection.} 

Unlike instance-level contrast that only draws support from fine-grained information at the bottom partition, cluster-wise contrast takes both coarse-grained and fine-grained semantics at multiple partitions into account. Such approaches~\cite{pcl,swav} usually construct one positive pair consisting of the anchor and the prototype of the cluster it belongs to, while all remaining prototypes are treated as negative candidates, causing fake negatives. Our downward masking strategy is proposed to mask false negatives in the lower structure according to the higher-level semantic information brought by the upper structure of the hierarchical tree. In other words, during the selection process of contrastive pairs, MHCCL filters out false negative candidates that belong to the same cluster at next partition according to the hierarchy, and supplements them to the set of positive ones at current partition. As illustrated in Fig.~\ref{fig:overview}, in addition to the positive pair formed by $\mathbf{c}_{1}^{(1)}$ and $\mathbf{z}_{1}$, $\mathbf{c}_{2}^{(1)}$ and $\mathbf{c}_{1}^{(2)}$ are also treated as positive candidates, which are marked with red boxes. They are mistakenly regarded as negative candidates in previous approaches but they actually share similar higher-level semantics since they belong to the same cluster with the prototype of $\mathbf{c}_1^{(2)}$ at next partition $\text{P}^{(2)}$. To this end, MHCCL utilizes the hierarchical structure information from upper partitions to lower partitions, to filter out false negatives by the novel downward masking strategy. We treat clusters that have the same cluster label at next partition as positive candidate clusters, while those that do not belong to the same cluster as negative candidate clusters. Therefore, in addition to the unique positive pair formed by the anchor and its corresponding prototype, MHCCL also supplements $H_{\text{pos}}-1$ prototypes from positive candidate clusters. Meanwhile, the negative pairs are formed by the anchor and randomly selected $H_{\text{neg}}$ prototypes from negative candidate clusters. In this way, we obtain $H_{\text{pos}}$ positive and $H_{\text{neg}}$ negative cluster-wise pairs at each partition.

\subsection{Multi-Level Contrastive Loss}
In previous studies, the contrastive learning loss usually adopts a softmax activation function and a cross-entropy loss for a single classification task. The labels are usually one-hot vectors, as there is only one positive pair. In our work, the labels are multi-hot vectors consisting of a specified number of multiple positive and negative samples in instance-wise or cluster-wise contrastive learning. So we treat the task as a multi-label classification problem, which is achieved by the sigmoid activation function and multiple binary cross-entropy losses. The overall loss is calculated by summing up multiple single binary cross-entropy losses. The multi-level contrastive loss consisting of the instance-wise and cluster-wise loss is calculated as follows:

\paragraph{Instance-Wise Contrastive Loss.}
Eq.~\eqref{instanceloss} defines the instance-wise contrastive loss function $\mathcal{L}_{\text{ins}}^{~(i)}$, where $\sigma(\cdot)$ is the sigmoid activation function. $\mathds{1}_{[1 \leqslant j\leqslant 2 \times S_{\text{pos}}]} \in \{0,1\} $ and $\mathds{1}_{[ 2 \times S_{\text{pos}} +1 \leqslant j\leqslant  2 \times S_{\text{all}}]} \in \{0,1\}$ are all indicator functions. $\text{sim}(\mathbf{z}_{i}^{a},\mathbf{u}_{i,j})$ computes the relevance score between the anchor $\mathbf{z}_{i}^{a}$ and each selected instance $\mathbf{u}_{i,j}$ via cosine similarity. The selected instance $\mathbf{u}_{i,j}$ is defined in Eq.~\eqref{uij}, where $a$ and $b$ denote two different augmented views of each instance. $S_{\text{all}} = 2 \times S_{\text{pos}}+2 \times S_{\text{neg}}$ is the total number of selected instance-wise contrastive pairs, where $S_{\text{pos}}$ and $S_{\text{neg}}$ are the number of positive and negative instances respectively. The number is multiplied by $2$ because each instance has two augmented views that can be used for contrastive learning,

\begin{equation}
\begin{aligned}
\mathcal{L}_{\text{ins}}^{~(i)}=-\sum_{j=1}^{S_{\text{all}}}[\mathds{1}_{[1 \leqslant j\leqslant 2 \times S_{\text{pos}} ]}\cdot \log\sigma (\text{sim}(\mathbf{z}_{i}^{a},\mathbf{u}_{i,j}))\\
+\mathds{1}_{[ 2 \times S_{\text{pos}} +1 \leqslant j\leqslant  2 \times S_{\text{all}}]}\cdot \log(1-\sigma (\text{sim}(\mathbf{z}_{i}^{a},\mathbf{u}_{i,j})))] ,\\
\end{aligned}
\label{instanceloss}
\end{equation}

\begin{equation}
\mathbf{u}_{i,j} = \left\{
\begin{array}{rl}
\mathbf{z}_{i,j}^{a}, & \text{if } j\neq 1 ~\text{and}  ~\text{mod}(j, 2) \neq 0 .\\
\mathbf{h}_{i,j}^{b}, & \text{otherwise }.
\end{array} \right.
\label{uij}
\end{equation}

\paragraph{Cluster-Wise Contrastive Loss.}
Since cluster-wise contrastive learning utilizes the hierarchical structure, we calculate the loss at each partition and then take the average loss of $M$ partitions as the final result. Eq.~\eqref{clusterloss} defines the cluster-wise contrastive loss function $\mathcal{L}_{\text{clu}}^{~(i)}$. The definitions of indicator functions are similar to those in instance-wise contrastive loss. $H_{\text{all}} = H_{\text{pos}}+H_{\text{neg}}$ is the total number of cluster-wise contrastive pairs, where $H_{\text{pos}}$ and $H_{\text{neg}}$ are the number of selected positive and negative prototypes respectively. $\mathbf{v}_{i,j}$ is randomly selected from the set of recalculated prototypes $C=\{\mathbf{c}_k^p\}_{k=1}^{K_p}$ at multiple partitions, where $K_p$ is the number of clusters at the $p$-th partition,

\begin{equation}
\begin{aligned}
\mathcal{L}_{\text{clu}}^{~(i)}=-\sum_{j=1}^{H_{\text{all}}}\frac{1}{M}\sum_{p=1}^{M}[\mathds{1}_{[1 \leqslant j\leqslant H_{\text{pos}} ]}\cdot \log\sigma (\text{sim}(\mathbf{z}_{i}^{a},\mathbf{v}_{i,j}))\\
+\mathds{1}_{[ H_{\text{pos}} +1 \leqslant j\leqslant H_{\text{all}}]}\cdot \log(1-\sigma (\text{sim}(\mathbf{z}_{i}^{a},\mathbf{v}_{i,j})))].
\end{aligned}
\label{clusterloss}
\end{equation}

\paragraph{Overall Loss.} The overall multi-level contrastive loss is finally obtained by summing up the instance-wise loss and the cluster-wise loss of all $B$ instances in the minibatch, which is defined as $\mathcal{L}_{\text{overall}}=\sum_{i=1}^{B}(\mathcal{L}_{\text{ins}}^{~(i)}+\mathcal{L}_{\text{clu}}^{~(i)})$. 

\begin{table}[t]
    \centering
    \setlength{\tabcolsep}{10pt}
    \renewcommand{\arraystretch}{0.95}
    \begin{tabular}{c|cccc}
    \toprule[1.1pt]
        Dataset  & $N$ & $V$ & Classes & Length   \\
        \midrule
        HAR & 10,299 & 9 & 6 & 128  \\
        WISDM  & 4,091 & 3 & 6 & 200  \\ 
        SHAR & 11,771 & 3 & 17 & 151  \\ 
        Epilepsy & 12,500 & 1 & 2 & 178 \\
        PenDigits & 10,992 & 2 & 10 & 8  \\ 
        EW  & 259 & 6 & 5 & 17,984  \\ 
        FM & 416 & 28 & 2 & 50  \\ 
        \bottomrule[1.1pt]
    \end{tabular}
    \caption{Statistical information of 7 datasets. $N$ denotes the number of instances and $V$ denotes the number of variables.}
    \label{table:dataset}
\end{table}

\section{Experiments}

\subsection{Experimental Settings}

\paragraph{Datasets.} We select 7 widely-used multivariate time series datasets from popular archives after taking into consideration of the application fields, length of sequences, number of instances, etc. The datasets include SHAR~\cite{shar}, Epilepsy~\cite{epilepsy}, WISDM~\cite{wisdm}, HAR from UCI\footnote{http://archive.ics.uci.edu/ml/datasets.php} archive, and PenDigits, EW (EigenWorms), FM (FingerMovements) from the UEA\footnote{http://www.timeseriesclassification.com/dataset.php} archive~\cite{dtw}. More details are listed in Table~\ref{table:dataset}. We split the data into 80\% and 20\% for training and testing, and use 20\% of the training data for validation in datasets except those from UEA archive which go through a pre-defined train-test split.

\begin{table*}[htbp]
  \centering
   \small
   \setlength{\tabcolsep}{3.4pt}
   \renewcommand{\arraystretch}{0.93}
    \begin{tabular}{c|c|ccccccccc}
    \toprule[1.1pt]
    Dataset & Metric & SimCLR & BYOL  & T-Loss & TSTCC & TS2Vec & SwAV  & PCL   & CCL   & MHCCL \\ 
    \midrule
    \multirow{3}[1]{*}{HAR} & ACC   & 81.06±2.35 & 89.46±0.17 & 91.06±0.94 & 89.22±0.70 & 90.47±0.66 & 68.81±1.50 &   74.49±1.95    &   86.84±1.08    & \textbf{91.60±1.06} \\
          & MF1   & 80.62±2.31 & 89.31±0.17 & 90.94±0.96 & 89.23±0.76 & 90.46±0.64 & 66.69±1.56 &   65.02±1.69    &   83.56±0.89    & \textbf{91.77±1.11} \\
          & $\kappa$ & 77.25±2.82 & 87.33±0.20 & 89.26±1.13 & 87.03±0.85 & 89.15±0.79 & 62.41±1.81 &   71.96±2.12    &    81.46±1.47   & \textbf{89.90±1.27} \\
    \midrule
    \multirow{3}[0]{*}{WISDM} & ACC   & 83.04±4.21 & 87.84±0.38 & 91.48±1.05 & 81.48±0.98 & 92.33±1.05 & 73.44±1.28 & 69.47±1.51 &  85.18±1.23     & \textbf{93.60±1.06} \\
          & MF1   & 75.83±6.47 & 84.02±0.77 & 88.79±1.53 & 69.17±2.25 & 90.27±1.07 & 53.90±3.56 & 61.75±3.29 &  81.22±1.19     & \textbf{91.70±1.10} \\
          & $\kappa$ & 75.15±6.27 & 82.43±0.57 & 87.79±1.51 & 73.13±1.33 & 90.36±1.49 & 59.75±2.29 & 56.36±2.57 &  79.19±1.28     & \textbf{90.96±1.08} \\
    \midrule
    \multirow{3}[0]{*}{SHAR} & ACC   & 67.22±1.76 & 67.00±0.98 & 80.88±3.94 & 70.80±0.75 & 82.94±2.91 & 57.49±2.75 & 56.28±1.47      &   75.13±2.21   & \textbf{83.42±1.76} \\
          & MF1   & 53.05±2.42 & 59.30±1.30 & 77.06±3.44 & 64.28±0.72 & 77.89±2.95 & 58.82±2.53 &   49.42±1.62    &   64.77±2.47    & \textbf{78.45±2.09} \\
          & $\kappa$ & 63.84±1.97 & 63.62±1.10 & 79.65±4.35 & 67.74±0.85 & 78.94±3.20 & 58.89±3.50 &   51.78±1.65    &    72.65±1.32   & \textbf{80.58±1.84} \\
    \midrule
    \multirow{3}[0]{*}{Epilepsy} & ACC   & 93.00±0.57 & \textbf{98.08±0.09} & 96.94±0.20 & 97.19±0.18 & 96.32±0.23 & 94.30±0.85 &   89.93±1.34    &   95.47±0.88    & 97.85±0.49\\
          & MF1   & 88.09±0.97 & \textbf{96.99±0.15} & 95.20±0.30 & 95.47±0.31 & 94.27±0.37 & 90.80±1.37 &   87.68±1.42    &   91.38±1.60    & 95.44±0.82 \\
          & $\kappa$ & 76.27±1.93 & \textbf{93.99±0.30} & 90.41±0.61 & 90.94±0.61 & 88.54±0.74 & 81.62±2.73 &   85.78±1.88    &   79.42±1.23    & 91.08±1.57 \\
    \midrule
    \multirow{3}[0]{*}{PenDigits} & ACC   & 93.35±0.17 & 94.93±0.08 &  97.86±0.52  & 97.44±0.23 &  97.83±0.24 & 78.02±2.86 &    86.18±1.25   &    91.27±1.46   & \textbf{98.69±0.41} \\
          & MF1   & 93.27±0.17 & 94.96±0.07 & 97.87±0.52 & 97.45±0.23 & 97.80±0.25 & 76.86±3.06 &   83.36±1.11    &  88.61±0.96     & \textbf{98.71±0.55} \\
          & $\kappa$ & 92.61±0.19 & 94.37±0.09 & 97.63±0.57 & 97.16±0.26  & 97.59±0.27 & 71.10±3.19  &  80.69±1.01     &  87.66±0.61     & \textbf{97.43±0.72} \\
    \midrule
    \multirow{3}[0]{*}{EW} & ACC   & 60.87±1.84 & 79.39±0.84 & 75.00±3.39 & 73.21±1.44 & \textbf{82.80±2.52} & 43.75±1.80 &  43.28±1.06     & 73.13±0.31 & 79.10±2.21 \\
          & MF1   & 55.62±4.69 & 75.12±1.05 &  67.85±3.64 & 64.33±0.81 & \textbf{81.59±3.73} & 35.16±2.31 &  40.02±2.01 &      65.49±1.57 & 76.01±1.45 \\
          & $\kappa$ & 47.99±3.27 & 71.84±1.19 & 65.12±4.91 & 63.41±1.17 & \textbf{80.74±3.41} & 21.82±2.27 &   22.58±1.59 &       63.32±2.44 & 69.53±1.90 \\ 
    \midrule
    \multirow{3}[0]{*}{FM} & ACC   & 49.20±0.98 & 49.60±1.11 & 50.50±1.72 & 50.17±1.60 & 50.00±1.63 & 44.22±2.19 &  50.44±1.95     &   50.23±1.12     & \textbf{52.09±1.45} \\ 
          & MF1   & 43.08±4.90 & 49.38±1.12 & 50.33±1.76 & 49.07±2.10 & 49.99±1.64 & 44.19±2.20 &  43.68±1.30     &    47.32±1.02  & \textbf{50.51±2.06} \\
          & $\kappa$ & -0.69±2.29 & -0.49±2.22 & 4.01±6.65 & 0.33±4.58 & -0.01±3.30 & 8.47±4.40 &   1.19±3.89    &   9.18±6.55    &  \textbf{17.87±3.39}\\ 
    \bottomrule[1.1pt]
    \end{tabular}
    \caption{Comparisons between our proposed MHCCL model against baselines on 7 datasets.}
  \label{tab:results}
\end{table*}

\paragraph{Evaluation Metrics.}
We choose linear classification as the downstream task to verify the quality of representations learned by different models. We adopt the following three widely-used metrics, i.e., accuracy ACC $= \frac{\text{TP}+\text{TN}}{\text{TP}+\text{TN}+\text{FP}+\text{FN}}$, macro-averaged F1-score MF1 $=\frac{2\times \text{PR}}{\text{P}+\text{R}}$ and Cohen's Kappa coefficient $\kappa = \frac{\text{ACC}-p_e}{1-p_e}$, where TP, TN, FP, FN are true positive, true negative, false positive and false negative respectively. Precision, recall and the hypothetical probability of chance agreement $p_e$ are calculated by $\text{P} =\frac{\text{TP}}{\text{TP}+\text{FP}}, \text{R} =\frac{\text{TP}}{\text{TP}+\text{FN}}$, and $p_e=\frac{[(\text{TP}+\text{FN}) *(\text{TP}+\text{FP})+(\text{FP}+\text{TN})*(\text{FN}+\text{TN})]}{ N^{2}}$, where $N$ denotes the total number of instances. For approaches with the hierarchical structure consisting of multiple partitions, the averaged performance at all partitions is reported.

\paragraph{Implementation Details.}
All models are implemented with PyTorch and the experimental evaluations are conducted on an NVIDIA GeForce RTX 3090 GPU. The data augmentation method is fixed to a combination of the weak (i.e. jitter-and-scale) and strong augmentation (i.e. permutation-and-jitter) in agreement with~\cite{tstcc}. ResNet~\cite{he2016deep} is adopted as the backbone network. The output dimension of the backbone is 128. The batch size $B$ is 128 by default and is reduced to 16 or 32 for small datasets. Stochastic Gradient Descent (SGD) is adopted as the optimizer and each model is trained for 200 epochs. In addition, in order to compare with algorithms that need to pre-specify the number of clusters such as K-Means, we assign them the actual number of classes of each dataset. We freeze the representations pretrained by all unsupervised learning models, and attach a linear classifier on top of them. We run each experiment five times and report the mean and standard deviation of the results.

\subsection{Results and Analysis}
\paragraph{Comparisons with Baseline Approaches.}
We compare MHCCL with eight state-of-the-art approaches in two categories: 1) instance-wise approaches including SimCLR~\cite{simclr}, BYOL~\cite{byol}, T-Loss~\cite{tloss}, TSTCC~\cite{tstcc} and TS2Vec~\cite{ts2vec}, and 2) cluster-wise approaches including SwAV~\cite{swav}, PCL~\cite{pcl} and CCL~\cite{ccl}. The overall results are listed in Table~\ref{tab:results}. In general, our proposed MHCCL achieves the best in 5 out of 7 datasets, especially compared to previous cluster-wise approaches including SwAV, PCL and CCL. Such performance gain effectively indicates that MHCCL is able to recognize and reduce false pairs by incorporating hierarchical clustering results at different granularities. The results show that instance-wise approaches achieve better results than cluster-wise approaches based on K-Means clustering (i.e. SwAV and PCL). It can be inferred that flat clustering which only captures a single hierarchy of semantic clusters is usually not enough. Because there is no guarantee to make sure the results obtained by each clustering are always correct, where false negatives will affect the discrimination between instances. Notably, MHCCL performs better in datasets with more classes such as SHAR, as the hierarchical structure in such dataset is more obvious. TS2Vec also performs well which incorporates the hierarchical information into learning. Unlike TS2Vec which focuses on the hierarchy within the same time series from timestamps to the complete sequence, MHCCL focuses on hierarchical natural groupings behind the entire set of time series.

\paragraph{Effects of Specifying the Number of Clusters.}
As shown in Table~\ref{tab:specifying}, we compare our approach with its counterpart version which specifies the number of clusters $K$ at the top partition. Overall, the natural bottom-up clustering process is more beneficial than specifying $K$ in advance. The performance drops a lot on most datasets, while only those with smaller $K$ such as Epilepsy and FingerMovements are relatively less affected. This may be attributed to the fact that once the number of clusters in a certain partition is found to be less than $K$, the algorithm will fall back to the previous partition and merge clusters forcefully, which is conflicted with natural groupings in the underlying structure of data.

\begin{table}[!ht]
    \centering
    \setlength{\tabcolsep}{10pt}
    \renewcommand{\arraystretch}{0.9}
    \begin{tabular}{c|c|c}
    \toprule[1.1pt]
        Dataset  & w/o specifying $K$ & specifying $K$ \\ 
    \midrule
        HAR  &91.60±1.06 & 89.32±0.73 \\ 
        WISDM & 93.60±1.06  & 89.11±1.81  \\ 
        SHAR  &83.42±1.76 & 77.69±1.64  \\ 
        Epilepsy &97.85±0.49 &  96.42±0.34 \\ 
        PenDigits &98.69±0.41 & 91.76±0.89\\
        EW &79.10±2.21 & 73.28±1.91 \\ 
        FM &52.09±1.45 & 50.32±1.37  \\ 
    \bottomrule[1.1pt]
    \end{tabular}
    \caption{Effects of specifying number of clusters in ACC.}
  \label{tab:specifying}
\end{table}

\begin{table}[htbp]
  \centering
  \small
    \begin{tabular}{l|ccc}
    \toprule[1.1pt]
    Partitions & ACC & MF1 & $\kappa$ \\
    \midrule
     $\text{P}^{(1)}$ &   90.59±0.74 & 87.76±1.21 & 85.66±1.02  \\ 
     $\text{P}^{(2)}$  & 87.36±0.40 & 83.62±0.95 & 81.11±0.54  \\
     $\text{P}^{(3)}$  & 89.21±1.20 & 86.44±1.94 & 83.72±1.68  \\
     \midrule
     $\text{P}^{(1)}+\text{P}^{(2)}$ & 90.97±0.63 & 87.16±0.74 & 84.19±0.84  \\
    $\text{P}^{(1)}+\text{P}^{(3)}$ & 91.15±0.87 & 90.19±1.77 & 89.70±0.93  \\
    $\text{P}^{(2)}+\text{P}^{(3)}$ & 90.22±1.69 & 87.45±1.72 & 85.32±1.41  \\ 
     \midrule
    $\text{P}^{(1)}+\text{P}^{(2)}+\text{P}^{(3)}$ & 93.60±1.06 & 91.70±1.10 & 90.96±1.08  \\ 
    \bottomrule[1.1pt]
    \end{tabular}
    \caption{Effects of utilizing different partitions in downward masking to select cluster-level contrastive pairs on WISDM.}
    \label{tab:partitions}
\end{table}

\paragraph{Effects of Using Different Partitions in Downward Masking.}
To evaluate the impact of clustering at different granularities on performance, we test the results of using different combinations of partitions to select cluster-level contrastive pairs in MHCCL. As shown in Table~\ref{tab:partitions}, the bottom partition $\text{P}^{(1)}$ plays a more important role than the upper ones such as $\text{P}^{(2)}$ and $\text{P}^{(3)}$ in downward masking. This confirms our intuition that the lower partition preserves fine-grained information with higher purity. Meanwhile, the coarse-grained information brought by the top partition is also not negligible, as each cluster consists of a relatively large number of instances, which reflects higher-level latent semantics. It can be observed that the combination of all partitions works the best for downward masking, confirming our motivation.

\paragraph{Effects of Using Different Upward Masking Strategies.}
In addition to masking outliers with thresholds, we also investigate two more strategies, including masking a fixed proportion of outliers and replacing each calculated prototype with the original instance closest to it. The results evaluated on WISDM dataset are summarized in Table~\ref{tab:masking}. It can be observed that masking instances with a proper threshold of $0.3$ at the bottom clustering partition achieves the highest accuracy. 
When the threshold is either too small or too large, the results can be reduced. This may be attributed to the fact that the number of masked instances plays an important role on the precision of prototype recalculation, which then affects the selection of positive and negative candidates especially in cluster-wise contrastive learning. Removing outliers proportionally also achieves effective results. Overall, masking outliers at the bottom partition performs better than masking them at multiple partitions. In addition, replacing the calculated mean prototypes with the original time series also helps improve the performance to some extent, but it is still slightly inferior compared to direct masking strategies.

\begin{table}[htbp]
\centering
\small
\renewcommand{\arraystretch}{0.96}
 \setlength{\tabcolsep}{5pt}
    \begin{tabular}{l|l|c|c}
    \toprule[1.1pt]
    Masking Strategy   & Partitions     & Parameter & ACC \\
\midrule
\multirow{5}[0]{*}{mask\_threshold}    & $\text{P}^{(1)}$           & 0.1   & 91.26±0.23   \\
                   & $\text{P}^{(1)}$           & 0.3   & 93.60±1.06    \\
                   & $\text{P}^{(1)}$          & 0.9   & 90.99±0.05   \\
                   & $\text{P}^{(2)}$+$\text{P}^{(3)}$      & 0.3   & 88.47±1.20   \\
                   & $\text{P}^{(1)}$+$\text{P}^{(2)}$+$\text{P}^{(3)}$ & 0.3   & 90.62±1.49   \\
\midrule
\multirow{5}[0]{*}{mask\_proportion}   & $\text{P}^{(1)}$           & 10\%  & 91.48±0.05   \\
                   & $\text{P}^{(1)}$           & 50\%  & 88.30±0.43    \\
                   & $\text{P}^{(1)}$           & 70\%  & 76.78±0.39   \\
                   & $\text{P}^{(2)}$+$\text{P}^{(3)}$      & 10\%  & 88.65±1.09   \\
                   & $\text{P}^{(1)}$+$\text{P}^{(2)}$+$\text{P}^{(3)}$ & 10\%  & 89.41±0.24   \\
\midrule
\multirow{3}[0]{*}{replace\_prototypes} & $\text{P}^{(1)}$           & -     & 91.11±0.09   \\
                   & $\text{P}^{(2)}$+$\text{P}^{(3)}$      & -     & 89.04±0.09   \\
                   & $\text{P}^{(1)}$+$\text{P}^{(2)}$+$\text{P}^{(3)}$ & -     & 89.53±1.16  \\
    \bottomrule[1.1pt]
    \end{tabular}
    \caption{Effects of different upward masking strategies with different parameters in hierarchical clustering on WISDM.}
    \label{tab:masking}
\end{table}

\paragraph{Ablation Analysis. }
In order to evaluate the validity of each proposed module in MHCCL, Table~\ref{tab:ablation} presents comparisons between MHCCL and its five variants on WISDM dataset when certain strategies are removed. Overall, each module of MHCCL plays an essential role in our task. The results show that the variant without hierarchical clustering performs the worst, which confirms that our motivation of incorporating multi-granularity clustering results from the hierarchy is effective. In addition, it can be observed that the cluster-level contrast plays a more important role than the instance-level contrast, which may be attributed to the good performance of applying upward and downward masking strategies in cluster-wise contrastive pairs construction. Furthermore, the downward masking strategy achieves better results than the upward one, which highlights the importance of exploiting latent higher-level semantics in the clustering hierarchy. To be specific, according to ground-truth labels, MHCCL with downward masking helps reduce 27.30\% instance-wise false negative pairs and 37.29\% cluster-wise ones compared to flat K-Means clustering which specifies $K=6$ on WISDM dataset.

\begin{table}
  \centering
  \small 
  \renewcommand{\arraystretch}{0.96}
  \setlength{\tabcolsep}{2.5pt}
    \begin{tabular}{l|ccc}
    \toprule[1.1pt]
         ~ & ACC & MF1 & $\kappa$  \\
    \midrule
    MHCCL & 93.60±1.06 & 91.70±1.10 & 90.96±1.08      \\
    w/o cluster-level contrast & 89.67±0.71 & 88.03±0.79 &	87.04±1.29 \\
    w/o instance-level contrast &    91.58±1.11 &	89.75±1.26 &	88.10±1.32    \\
    w/o hierarchical clustering &    83.96±0.53 &	80.19±1.28	& 77.44±0.75    \\
    w/o downward masking & 87.83±0.64 & 84.89±0.98 &  82.84±0.93 \\
    w/o upward masking & 90.05±1.17 & 88.67±1.05	& 87.49±1.37	 \\
    \bottomrule[1.1pt]
    \end{tabular}
    \caption{Ablation results on WISDM dataset.}
  \label{tab:ablation}
\end{table}

\section{Conclusion}  
In this paper, we propose a \underline{M}asked \underline{H}ierarchical \underline{C}luster-wise \underline{C}ontrastive \underline{L}earning (MHCCL) model for multivariate time series. Motivated by the observation that fine-grained clustering preserves higher purity, and coarse-grained clustering reflects higher-level semantics, MHCCL proposes to utilize such information for better constructions of pairs in contrastive learning. Specifically, the upward masking strategy removes outliers while updating prototypes iteratively with hierarchical clustering. The downward masking strategy guides the selection of cluster-wise contrastive pairs. In this way, the latent positive pairs can be mined while the fake negative pairs can be filtered out. The conducted experiments show that MHCCL effectively benefits the downstream time series classification tasks.

\section{Acknowledgments}
This research is partially supported by the National Key R\&D Program of China No.2021YFF0900800; the NSFC No.91846205; Shandong Provincial Key Research and Development Program (Major Scientific and Technological Innovation Project) (NO.2021CXGC010108); Shandong Provincial Natural Science Foundation (No. ZR202111180007); the Fundamental Research Funds of Shandong University, and the State Scholarship Fund supported by the China Scholarship Council (CSC). H.Qian thanks the support from the Wallenberg-NTU Presidential
Postdoctoral Fellowship.

\end{document}